\documentclass[mnsc,nonblindrev]{informs3}

\OneAndAHalfSpacedXI

\usepackage{amsmath,amssymb,amsfonts,algorithm,algorithmic,bm}

\usepackage{natbib}
 \bibpunct[, ]{(}{)}{,}{a}{}{,}%
 \def\bibfont{\small}%

\usepackage{rotating}
\usepackage{fancyvrb,makecell}

\TheoremsNumberedThrough     
\ECRepeatTheorems
\JOURNAL{Any INFORMS Journal}

\EquationsNumberedThrough    

\MANUSCRIPTNO{IJDS-0001-1922.65}

\begin{document}


\RUNTITLE{}

\TITLE{Daily Physical Activity Monitoring---Adaptive Learning from Multi-source Motion Sensor Data}


\ARTICLEAUTHORS{%
\AUTHOR{Haoting Zhang\thanks{equal contribution}}

\AFF{University of California, Berkeley, \EMAIL{haoting\_zhang@berkeley.edu}}

\AUTHOR{Donglin Zhan\footnotemark[1]}

\AFF{Columbia University, \EMAIL{dz2478@columbia.edu}}

\AUTHOR{Yunduan Lin, Jinghai He}

\AFF{University of California, Berkeley, \EMAIL{yunduan\_lin,jinghai\_he@berkeley.edu}}

\AUTHOR{Qing Zhu}
\AFF{Lawrence Berkeley National Laboratory, \EMAIL{qzhu@lbl.gov}}

\AUTHOR{Zuo-Jun Max Shen, Zeyu Zheng}
\AFF{University of California, Berkeley, \EMAIL{maxshen,zyzheng@berkeley.edu}}

}

\ABSTRACT{%
In healthcare applications, there is a growing need to develop machine learning models that use data from a single source, such as that from a wrist wearable device, to monitor physical activities, assess health risks, and provide immediate health recommendations or interventions. However, the limitation of using single-source data often compromises the model's accuracy, as it fails to capture the full scope of human activities. While a more comprehensive dataset can be gathered in a lab setting using multiple sensors attached to various body parts, this approach is not practical for everyday use due to the impracticality of wearing multiple sensors. To address this challenge, we introduce a transfer learning framework that optimizes machine learning models for everyday applications by leveraging multi-source data collected in a laboratory setting. We introduce a novel metric to leverage the inherent relationship between these multiple data sources, as they are all paired to capture aspects of the same physical activity. Through numerical experiments, our framework outperforms existing methods in classification accuracy and robustness to noise, offering a promising avenue for the enhancement of daily activity monitoring.

}

\maketitle

\section{Introduction}

The healthcare sector is undergoing a transformative era, fueled by the integration of artificial intelligence, data analytics, and sensor technology \citep{lee2022real,pillai2023rare}. As a part of this transformation, wearable motion sensor technologies are emerging as a key driver in reshaping personal health monitoring. These devices enable continuous and non-intrusive tracking of physical activities, providing invaluable insights into individual daily routines \citep{liu2021wearable}. For instance, for office workers, these sensors serve as proactive health partners, encouraging individuals to cultivate healthier habits and mitigating the risks associated with sedentary lifestyles \citep{adjerid2022gain}. Moreover, they offer potential life-saving benefits, such as monitoring vulnerable populations for sudden falls and triggering immediate alerts, thereby reducing the likelihood of severe injuries or complications \citep{chander2020wearable}.

Thanks to the increased accessibility of data and advancements in machine learning techniques, sensors can now precisely capture activities based on the collected data \citep{pandl2021trustworthy,xu2023multi,chen2023missing,matton2023contrastive}. This data typically comprises location trajectory information, capturing the movements of different body parts during specific activities. For thorough data acquisition, a laboratory setting is essential where multiple wearable motion sensors can be strategically positioned on various parts of an individual's body, such as the wrists, ankles, chest, and head. Participants are then asked to execute specific actions, which serve as the labels for the corresponding data \citep{mccarthy2015motion}. Time series classification methods are then used to analyze and interpret this multi-source data, enabling the construction of prediction models \citep{zeng2020arm}. This comprehensive data collection method offers a holistic view of human movements, contributing to the high accuracy of the resulting models.

\subsection{Challenges}
However, in daily applications, it is often impractical or undesirable for users to wear multiple motion sensors. Instead, they prefer a single sensor placed on a particular body part, like the left arm, thus gathering data solely from that single source. This limitation poses a dual challenge. On one hand, models that have been trained with multi-source data can be challenging to adapt when provided with the limited perspective of single-source data in daily applications. On the other hand, training classifiers solely with data from one source can compromise activity detection accuracy, as it lacks the broad spectrum of insights that multi-source data offers. 

To bridge this gap, we advocate for the adoption of transfer learning techniques \citep{spathis2021self,merrill2023self}, enabling the effective utilization of multi-source data in daily monitoring tasks. These techniques draw upon the knowledge from interconnected datasets to accomplish the classification task. Moreover, a distinctive characteristic of the multi-source data collected by motion sensors is its inherent pairwise structure\textemdash a facet not fully exploited by existing transfer learning methodologies. As previously mentioned, the multi-source data emerges when participants wear an assortment of sensors across various body parts, engaging in specific predefined actions. This setup ensures that time series data from these domains are acquired synchronously, all marked with the same action labels. Such cohesive data collection naturally establishes an intrinsic data pairing between different domains, offering a rich set of correlations that could potentially improve the performance of classification outcomes. On the other hand, existing transfer learning methods fall short in this aspect, processing data from each sensor independently. This oversight may miss out on potential gains in achieving a more comprehensive understanding of activities.

\subsection{Proposed Approach \& Contribution}
\begin{figure}[htbp]
    \centering
    \includegraphics[width=\textwidth]{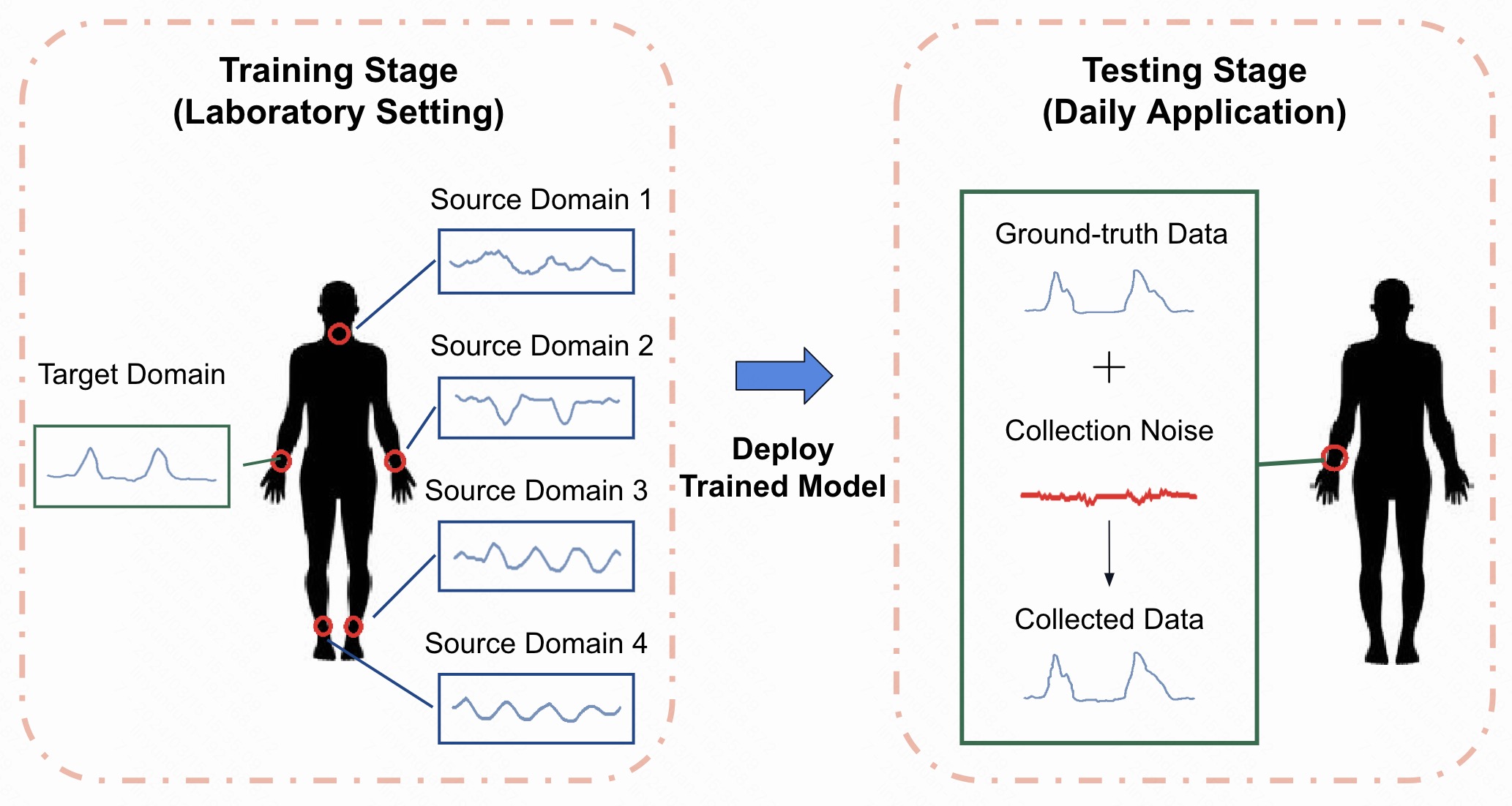}
    \caption{The procedure of training a classifier and applying the trained classifier in daily physical activity monitoring.}
    \label{fig:illustration}
\end{figure}

\begin{figure}[htbp]
    \centering
    \includegraphics[width=\textwidth]{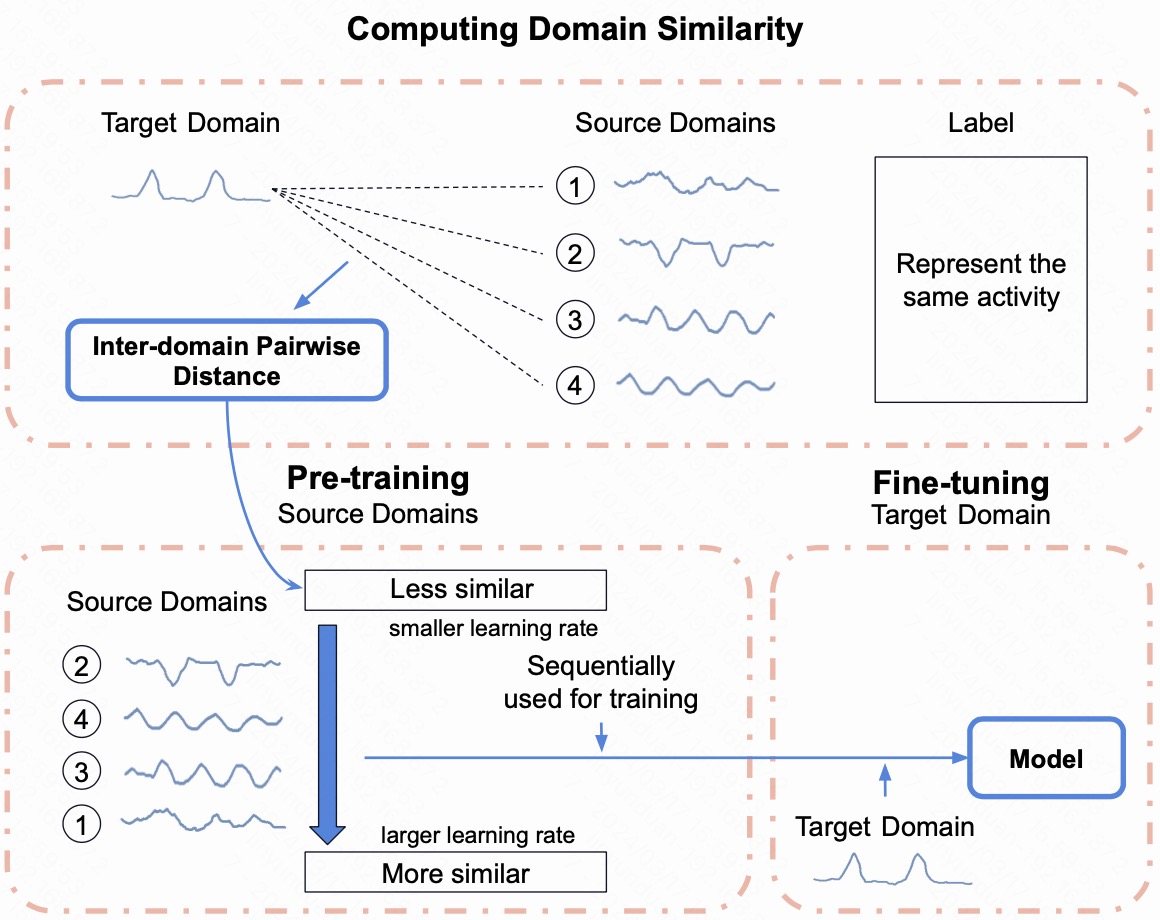}
    \caption{The illustration of the proposed transfer learning framework.}
    \label{fig:illustration-2}
\end{figure}

In this work, we introduce a novel transfer learning framework designed to harness the pairwise structure of multi-source motion sensor data. Within this framework, the sensor placed on the body part that is daily monitored acts as the target domain, while sensors on other body parts are considered as source domains. Our proposed framework includes three steps: (1) computing the domain similarities between the target domain and source domains; (2) pre-training the model on source domains based on the domains' similarities; and (3) fine-tuning the pre-trained model on the target domain. In the first step, we propose a novel metric named \textit{Inter-domain Pairwise Distance} (IPD), which factors in the pairwise structure. We pair time series data across different domains in a manner that reflects the data collection procedure and calculate IPD through the method of smooth bootstrapping. In the second step, we train the model (classifier) on all source domains, adjusting the step size (learning rate) based on the calculated IPD. A lower IPD value indicates a closer similarity between the source and target domains, prompting the model to adopt a larger step size and focus more on learning from these similar domains.
For the last step, we fine-tune the pre-trained model on the target domain, mirroring traditional transfer learning methods. 
The procedure of training a classifier and applying the trained classifier in daily physical activity monitoring as well as the proposed transfer learning framework is summarized in \textbf{Figures \ref{fig:illustration}} and \textbf{\ref{fig:illustration-2}}.

This work offers two main contributions to daily physical activity monitoring for healthcare applications. First, we propose a structured framework that leverages multi-source time series data collected in laboratory experiments into daily physical activity monitoring. Through our approach, laboratory-derived data is transformed, providing valuable insights applicable to real-world contexts. Secondly, we emphasize the unique pairwise structure inherent in motion sensor data, which sets it apart from many conventional transfer learning scenarios. 
We introduce a novel metric to reflect this structured pairing within the modeling framework. This metric has a wide applicability range, being compatible with various time series distance measures. 
Through rigorous testing on various datasets from the UCI Machine Learning Repository, we demonstrate that our novel framework outperforms existing state-of-the-art methods in both classification accuracy and robustness against input noise. These results mark it as a significant advancement in wearable sensor technology for healthcare applications.


\section{Related Work}
\label{sec:literature}
\paragraph{Physical Activity Monitoring in Healthcare}
Daily physical activity monitoring in healthcare has garnered significant attention in recent years due to the increasing prevalence of sedentary lifestyles and the associated risk of chronic diseases. In this context, wearable devices and smartphone-based sensors have emerged as popular tools for continuous monitoring of physical activity patterns. These devices typically employ accelerometers, gyroscopes, and heart rate monitors to track various aspects of physical activity, such as step count, energy expenditure, and activity type \citep{patel2015wearable,doherty2017large,merrill2023homekit2020}.

Researchers have explored the application of machine learning and deep learning algorithms to classify and predict different types of activities from sensor data \citep{hammerla2016deep,weatherhead2022learning}. These studies have demonstrated the potential of daily physical activity monitoring in improving the management of chronic conditions, such as diabetes, cardiovascular diseases, and obesity, by promoting patient adherence to prescribed exercise regimens and facilitating personalized treatment plans \citep{jakicic2016effect}. Moreover, the integration of physical activity monitoring with telemedicine and remote patient monitoring systems has enabled healthcare providers to remotely assess patients' progress and offer timely interventions \citep{kvedar2014connected,zheng2023machine}.

\paragraph{Time Series Classification}
Time series classification tasks have been significantly improved with the advent of deep learning techniques. Examples of these improvements include disease diagnosis based on time series of physiological parameters, the classification of heart arrhythmias from Electrocardiograms (ECGs) \citep{bagnall2017great}, and human activity recognition \citep{petitjean2014dynamic}. While certain deep learning architectures can achieve state-of-the-art results on large-scale data, their performance tends to be suboptimal when access to a large labeled training dataset is limited \citep{sutskever2014sequence,CHEN2021126}.

Owing to the challenges in collecting and annotating time-series data, researchers have increasingly opted against training large deep learning models from scratch. Instead, they employ pre-trained models on source tasks and adapt them to the target task. This transfer learning approach has yielded satisfactory model performance in a range of fields, including computer vision \citep{bhattacharjee2017identifying}, transportation \citep{sun2023meta,wang2023st}, anomaly detection \citep{bhattacharjee2017context,xiao2024xtsformer}, and data mining \citep{wang2022novel,zhan2024meta}, particularly when training data is limited. Prior studies have demonstrated that transfer learning can also enhance performance in time series classification problems with limited data \citep{fawaz2018transfer}.

\section{Problem Statement}
\label{sec:problem}

For effective daily physical monitoring, a model (classifier) is required to interpret the data collected by the daily wearable motion sensor, like smartwatches. This model is designed to accurately predict the type of physical activity being performed. In this context, we provide the mathematical definition of the \textbf{time series data} in Definition \ref{def:1}, which represents the data collected by a specific motion sensor.
\begin{definition}[Time series data]
\label{def:1}
Time series data is a sequence of observations collected sequentially over a time period. For a period $T$ of observations, the time series data is denoted as $X=\left[x_{1}, \ldots, x_{T}\right] \in \mathbb{R}^{K \times T}$, where each $x_{t}$ is a $K$-dimensional vector, representing the measurement taken at the $t$-th moment in the sequence.
\end{definition}

From Definition \ref{def:1}, time series data consist of a series of observations recorded by a motion sensor. Each observation in this series is represented by a $K$-dimensional vector, detailing motion attributes such as acceleration, rotation, and orientation at a particular timestamp. The entirety of this time series data $X$ represents a specific activity. This activity is categorized under the label $C$, which belongs to a collection of pre-determined activity types $\mathcal{C}$.



\begin{definition}[Domain]
\label{def:2}
    A domain $\mathcal{D}$ is comprised of a feature space $\mathcal{X}\subset\mathbb{R}^{K\times T}$ and a marginal distribution $P(\mathcal{X})$. Within domain $\mathcal{D}$, we have multiple samples of time series data $\hat{\mathcal{D}}=\left\{X_1,\ldots,X_N\right\}$ as a realization, where each $X_n\in \mathcal{X}$ denotes a time series data as defined in Definition \ref{def:1}. 
\end{definition}
 \begin{definition}[Pairwise multi-source time series]
\label{def:3}
A multi-source time series data $\mathfrak{X}\doteq\left\{\hat{\mathcal{D}}_1,\hat{\mathcal{D}}_2,\ldots,\hat{\mathcal{D}}_V\right\}$ is defined as a collection of time series data from multiple domains, where $\hat{\mathcal{D}}_v$ denotes the data set associated with the domain $\mathcal{D}_v$ defined in Definition \ref{def:2}. The pairwise structure of the multi-source time series data requires (i) each set $\hat{\mathcal{D}}_v$ has the same number of time series data (i.e., $N=\left |\hat{\mathcal{D}}_1   \right | =\left |\hat{\mathcal{D}}_2  \right |=\ldots=\left |\hat{\mathcal{D}}_V   \right |$), and (ii) for $n\in \left\{1,2,\ldots,N\right\}$, the $n$-th time series data across the domains $\left\{X_{\mathcal{D}_1,n},X_{\mathcal{D}_2,n},\ldots,X_{\mathcal{D}_V,n}\right\}$ are paired with an identical label $C_n\in \mathcal{C}$, where $X_{\mathcal{D}_v,n}$ denotes the $n$-th time series data in the domain $\mathcal{D}_v$. 
\end{definition}

In a laboratory setting, sensors placed on different body parts each define a unique domain, collectively gathering multi-source time series data to reflect specific actions performed by a participant. This necessitates aligning data from $V$ motion sensors across $V$ domains under consistent labels to represent the executed activities. Introducing a transfer learning framework aimed at 
leveraging lab-acquired, multi-domain time series data to improve wearable motion sensor data classification in real-world scenarios. By treating one sensor (the target domain $\mathcal{T}$) as the primary focus and the others (source domains $\mathcal{S}_q$, $q=1,\ldots,Q$, with $Q=V-1$) as supplementary, the framework learns a classifier from source domain data $\hat{\mathcal{S}_q} =\left\{X_{\mathcal{S}_q,n}\right\}_{n=1}^N$ before training it with target domain data $\hat{\mathcal{T}} =\left\{X_{\mathcal{T},n}\right\}_{n=1}^N$. This process integrates knowledge from source domains to enhance the classifier's ability to accurately categorize activities based on data from daily-used wearable sensors.


\section{Method}
\label{sec:method}
In this section, we present the adaptive transfer learning 
framework. Our method works as follows. In Section \ref{sec:41}, 
we introduce a new metric Inter-domain Pairwise Distance (IPD) to quantify the similarity between domains. We then provide a practical method to approximate IPD between each source domain and the target domain using the collected time series data. In Section \ref{sec:42}, we describe the procedure of pre-training the model within the source domains. The degree of knowledge transfer is reflected by adapting the learning rate 
based on the calculated IPD.
We postpone the description of the fine-tuning procedure in the target domain to Appendix.

\subsection{Domain Similarity Computation}
\label{sec:41}
In this section, we introduce the \emph{Inter-domain Pairwise Distance (IPD)}, a novel metric designed to quantify the similarity between two domains wherein the time series data share a pairwise structure.
\begin{definition}
\label{def:IPD}
Inter-domain Pairwise Distance (IPD) between two domains $\mathcal{S}$ and $\mathcal{T}$ is defined as 
\begin{equation*}
    \operatorname{IPD}= \mathbb{E}\left \{ \operatorname{dist} \left(X_{\mathcal{S}},X_{\mathcal{T}}\right) \right \},
\end{equation*}
where $X_{\mathcal{S}}$ and $X_{\mathcal{T}}$ are paired time series data from domains $\mathcal{S}$ and $\mathcal{T}$ respectively, and $\operatorname{dist}\left(\,\cdot\,,\,\cdot\,\right)$ is a selected distance measure for two time series data.
\end{definition}

As shown in Definition \ref{def:IPD}, the IPD measures the similarity between two domains by evaluating the expected distance between their paired time series data. A higher IPD value indicates less similarity between these two domains. Notably, our metric offers both flexibility and adaptability, as it can integrate any time series distance measure, including but not limited to, Euclidean distance, Minkowski distance, and dynamic time warping. Furthermore, our approach maintains the pairwise data structure. This contrasts with most of the transfer learning literature, where the pairwise structure is neither present nor considered. Instead, such methods typically treat each time series data point as an independent sample from the distribution within the domain and measure the distance between two domains using metrics for empirical distributions, such as the Wasserstein distance. Our experimental findings later highlight that disregarding the pairwise structure of motion sensor data and resorting to conventional domain distances can compromise classification accuracy.

It is worth noting that the calculation of IPD requires taking the expectation over domains, however in practice, we typically observe specific realizations. In the following, we describe a threefold procedure of estimating $\operatorname{IPD}$ given only a set of paired time series data: (i) Computation of empirical inter-domain difference. We start by calculating the empirical IPD using the available paired time series data samples. (ii) Difference density estimation. We then approximate the probability density function of the empirical IPD with the kernel density estimation \citep{silverman1986density}. (iii) Difference sampling and distance calculation. Finally, we generate new samples from this approximated density function and use these new samples to approximate $\operatorname{IPD}$. In essence, our methodology shares similar spirits with the smooth bootstrap method \citep{hall1989smoothing}. Compared with merely using the empirical $\operatorname{IPD}$, our approach inherits the advantages from the smooth bootstrap method, including robustness against noise, improved variability, and practicability with small-size samples. We now illustrate this threefold procedure in details.
 
\paragraph{Computation of Empirical Inter-domain Difference}

Recall that the time series data in the source domain $\mathcal{S}_q$ is denoted as $\hat{\mathcal{S}_q} =\left\{X_{\mathcal{S}_q,n}\right\}_{n=1}^N$ for $q\in [Q]$ and the data in the target domain is $\hat{\mathcal{T}} =\left\{X_{\mathcal{T},n}\right\}_{n=1}^N$. Since both $X_{\mathcal{S}_q,n}$ and $X_{\mathcal{T},n}$ are multivariate time series data, we first decompose them into $K$ univariate time series data as $X^{(k)}_{\mathcal{S}_q,n}$ and $X^{(k)}_{\mathcal{T},n}$ for $k\in [K]$, where $X^{(k)}$ denotes the $k$-th entry of the time series data $X$. In other words, this decomposition allows us to consider each type of movement information within the time series data separately. 

For each $n$, we compute the pairwise univariate time series distance $D^{(k)}_{q,n}=\operatorname{dist}\left(X^{(k)}_{\mathcal{S}_q,n},X^{(k)}_{\mathcal{T},n}\right)\in \mathbb{R}$. We then reintegrate all the univariate distances back into the vector form and obtain the difference vector associated with the $n$-th pair of multivariate time series data $X_{\mathcal{S}_q,n}$ and $X_{\mathcal{T},n}$ as $D_{q,n}=\left(D^{(1)}_{q,n},D^{(2)}_{q,n},\ldots,D^{(K)}_{q,n}\right)^{\top}$. Noted that the distances calculated in this study are exclusively between a specific source domain $\mathcal{S}_q$ and the target domain $\mathcal{T}$. Therefore, we keep $q$ in the subscript to distinguish among source domains, while we omit $\mathcal{S}$ and $\mathcal{T}$ for notational simplicity. Consequently, we use the set of differences between each pair of time series data $\mathcal{M}_q=\left\{D_{q,1},D_{q,2},\ldots,D_{q,N}\right\}$ to represent the empirical difference between source domain $\mathcal{S}_q$ and the target domain $\mathcal{T}$. We summarize the computation of empirical inter-domain difference in Algorithm \ref{Inter-domain Pairwise Distance}.

\begin{algorithm}[htbp]
\caption{Computation of Empirical Inter-domain Difference.}
\label{Inter-domain Pairwise Distance}
\begin{algorithmic}[1]
\STATE{\textbf REQUIRE:} Source domain data $\hat{\mathcal{S}}_q$, target domain data $\hat{\mathcal{T}}$.
\FOR{$n = 1,2,\ldots,N$}
\STATE Select the $n$-th pair of multivariate time series $(X_{\mathcal{S}_q,n},X_{\mathcal{T},n})$ in $\hat{\mathcal{S}}_q$ and $\hat{\mathcal{T}}$;
    \FOR{$k=1,2,\ldots, K$}
    \STATE Compute univariate time series distance as $D^{(k)}_{q,n}=\operatorname{dist}\left(X^{(k)}_{\mathcal{S}_q,n},X^{(k)}_{\mathcal{T},n}\right)\in \mathbb{R};$
    \ENDFOR
    \STATE Construct the difference vector of the $n$-th pair
    \begin{equation*}
    D_{q,n}=\left(D^{(1)}_{q,n},D^{(2)}_{q,n},\ldots,D^{(K)}_{q,n}\right)^{\top}\in\mathbb{R}^{K};
    \end{equation*}
\ENDFOR
\STATE Return $\mathcal{M}_q=\left\{D_{q,1},D_{q,2},\ldots,D_{q,N}\right\}.$
\end{algorithmic}
\label{Alg:1}
\end{algorithm}

We note that 
we opt to decompose the multivariate time series data into $K$ univariate time series data
in our approach. The reason is mainly two-fold. First, the realm of univariate time series has been extensively studied, resulting in a plethora of research on determining distances between such series. In contrast, generalizing these established distances to multivariate scenarios still remains a relatively uncharted domain. Second, it is essential to preserve the multivariate structure of domain similarity in the initial stages. By doing so, we ensure a holistic assessment of domain similarity, enriched by the pairwise structure in subsequent analytical steps. This not only furnishes a more nuanced insight but also sets a robust foundation for in-depth analyses.

\paragraph{Difference Density Estimation}
We here approximate the probability density function of the inter-domain difference between the source domain $\mathcal{S}_q$ and the target domain $\mathcal{T}$ with the attained sample set $\mathcal{M}_q=\left\{D_{q,1},D_{q,2},\ldots,D_{q,N}\right\}$, using kernel density estimation \citep{silverman1986density}:
\begin{equation}
\label{eq:kde}
\hat{Q}_q(D)=\frac{1}{N} \sum_{n=1}^{N} \mathcal{K}_{\mathbf{H}}\left(D-D_{q,n}\right),
\end{equation}
where the kernel density function $\mathcal{K}_{\mathbf{H}}(D)$ is defined as $\mathcal{K}_{\mathbf{H}}(D)=|\mathbf{H}|^{-1 / 2} \mathcal{K}\left(\mathbf{H}^{-1 / 2} D\right).$ Here, $\mathbf{H}$ is a selected symmetric and positive definite $K\times K$ matrix, referred to as the bandwidth. Meanwhile, $\mathcal{K}$ is the selected kernel function. In this work, we specifically select the Gaussian kernel function as a representative. That is, $\mathcal{K}_{\mathbf{H}}(D)=(2 \pi)^{-d / 2}\left|\mathbf{H}\right|^{-1 / 2} e^{-\frac{1}{2} D^{\top} \mathbf{H}^{-1} D}$.

\paragraph{Inter-domain Pairwise Distance Calculation}
 
In the last step to estimate the $\operatorname{IPD}$ between source domain $\mathcal{S}_q$ and the target domain $\mathcal{T}$, we first generate samples from the approximated probability density function (p.d.f.) of the inter-domain difference $\hat{Q}_q(D)$. Given this p.d.f., the samples can be efficiently generated by the Monte Carlo Markov chain (MCMC) algorithm \citep{asmussen2007stochastic}. Suppose we have $m$ generated samples from $\hat{Q}_q(D)$, denoted as $\left\{\hat{D}_{q,i}\right\}_{i=1}^m$. By placing all the generated samples into a matrix, we have $\hat{\bm{D}_q} = \left(\begin{array}{c}{\hat{D}_{q,1}}\ {\ldots} \ \hat{D}_{q,i}\ {\ldots} \ {\hat{D}_{q,m}} \end{array}\right)^{\top}\in \mathbb{R}^{K\times m}$. This matrix $\hat{\bm{D}}_q$ from sampling contains the difference information between the source domain $\mathcal{S}_q$ and the target domain $\mathcal{T}$. Consequently, we approximate $\operatorname{IPD}_q$ with the norm of the matrix $\hat{\bm{D}}_q$ as 
\begin{equation*}
\widehat{\operatorname{IPD}}_q\doteq g_q =\frac{1}{m}\left\|\hat{\bm{D}}_q\right\|.
\end{equation*}

For each source domain, the approximated IPD is attained following this procedure. We represent the IPD vector between all source domains and the target domain as $\bm{g}=\left(g_1,g_2,\ldots,g_Q\right)^{\top}\in \mathbb{R}^{Q}$. 

\subsection{Pre-training in Source Domains}
\label{sec:42}
In this section, we present the process of pre-training the model (classifier) in the source domains and describe how the estimated Inter-domain Pairwise Distance (IPD) guides the pre-training process. Our framework has two key aspects: (1) Unified model framework. We utilize a singular classifier model, which is sequentially trained and updated across all source domains. (2) Model Flexibility. The framework is inherently flexible, accommodating a diverse range of models including Long Short-Term Memory (LSTM), encoders, and others, without being constrained to a specific model type. We denote the selected model as 
\begin{equation}
\label{eq:model}
    \bm{f}(X;\bm{\theta}),
\end{equation}
where $\bm{\theta}$ denotes model parameters to be learned. The model takes a single-source time series data $X\in \mathbb{R}^{K\times T}$ as the input and outputs a label $C\in \mathcal{C}=\left\{c_1,c_2,\ldots,c_L\right\}$ for classification.

A pivotal aspect of our method is the utilization of the 
IPD vector $\bm{g}$, which adaptively adjusts the learning rate of the model within each source domain.
Specifically, we increase the learning rate for a source domain with a smaller IPD to enable the model to better incorporate information from that domain. This is because a larger learning rate results in a higher degree of knowledge transfer, as the model assimilates more information about the current source domain over the same number of learning epochs. Thus, the derived IPD serves as a proxy for the similarity between the source and target domains, guiding the model to transfer knowledge from the most relevant source domains. Regarding the sequence in which the source domains are processed during the pre-training process, we sort and renumber all the source domains in descending order based on the associated IPD $g_q$. Thus, the pre-trained model is updated by and relies more on the data in the source domains that are more similar to the target domain. 

In terms of the training process, we employ the gradient descent method to minimize the loss function. We sequentially perform the gradient descent steps on all the source domains $\left\{\mathcal{S}_q\right\}_{q=1}^Q$, maintaining a consistent initial learning rate, $\lambda^0$, and total learning epochs, $J$, across all these domains. After completing the $J$ learning epochs on source domain $\mathcal{S}_q$, the parameters acquired are used as the initial starting point for the subsequent source domain $\mathcal{S}_{q+1}$, continuing the training the model in domain $\mathcal{S}_{q+1}$. 

The $j$-th learning epoch on source domain $\mathcal{S}_q$ can be represented by
\begin{equation}\label{iterate:weight} 
    \boldsymbol{\theta}_{q}^{j+1} = \boldsymbol{\theta}_{q}^{j}-\lambda_{q}^{j} \nabla_{\boldsymbol{\theta}} \mathcal{J}_{\mathcal{S}_{q}}\left(\boldsymbol{\theta}_{q}^{j}\right), 
\end{equation}
where 
\begin{equation}\label{eq:loss-function}
\mathcal{J}_{\mathcal{S}_{q}}\left(\boldsymbol{\theta}_{q}^{j}\right)={\mathbb{E}}_{\left\{\left(X_{\mathcal{S}_q,n},C_n\right)\right\}_{n=1}^N}\mathcal{L}\left( \boldsymbol{\theta}_{q}^{j}\right)
\end{equation}
is the empirical loss function. $\mathcal{L}$ denotes the categorical cross-entropy loss function, which is commonly used for classification problems \citep{murphy2012machine}. Here, $\bm{\theta}_q^{j}$ is the learned parameter of the model (e.g., the weight parameters of a neural network) in the $j$-th learning epoch and $\lambda_q^{j}$ is the corresponding learning rate. Inspired by the adaptive learning rate decay for sequential training on domains \citep{mirzadeh2020understanding}, our adaptive transfer learning framework updates the learning rate $\lambda_{q}^{j}$ as
\begin{equation}\label{iterate:lr} 
   \lambda_{q}^{j+1}=\lambda_{q}^{j} \cdot \left( 1 - \alpha_{q}\right), 
\end{equation}
where $\alpha_{q}$ is the weight of source domain $\mathcal{S}_{q}$ relative to all source domains. It is normalized by the sum of all $Q$ importance values of all source domains as
\begin{equation}\label{importance-value}
    \alpha_{q} = \frac{g_{q}}{\sum_{l=1}^{Q}g_{l}}.
\end{equation}

In this manner, we quantitatively determine the degree of knowledge transfer from each source domain to the target domain. This ensures that the greater the similarity between a source domain and the target domain (indicated by a smaller $\alpha_q$), the more knowledge is transferred from that source domain
(achieved through a larger learning rate). 

We postpone the description of fine-tuning the model in the target domain to Appendix, and summarize our framework in Algorithm \ref{MSTL}. Once the learning procedure ends, the learned model parameter $\bm{\theta}_{\mathcal{T}}^{j}$ is used to represent the trained classifier $\bm{f}\left(X;\bm{\theta}_{\mathcal{T}}^{j}\right)$ as in \eqref{eq:model}. Consequently, when the daily wearable motion sensor collects new time-series data $X^*$, the trained classifier is then employed to classify the physical activity with $\bm{f}\left(X^*;\bm{\theta}_{\mathcal{T}}^{j}\right)$, which leverages the information provided by multi-source time series. 

\begin{algorithm}[htbp]
\caption{Adaptive Learning from Multi-source Motion Sensor Data.}
\begin{algorithmic}[1]
\STATE {\textbf REQUIRE:} Source domain data $\hat{\mathcal{S}}_q$, target domain data $\hat{\mathcal{T}}$, sequence of collected labels $\left\{C_n\right\}_{n=1}^N$, initial learning rate $\lambda^{0}$, number of learning epochs in each source domain $J$, number of learning epochs in target domain $J_\text{target}$, initial value of the model parameter $\bm{\theta}^{0}$, number of partitions $\bm{k}$, baseline learning rate in target domain $\lambda_{\mathcal{T}}$ and number of the maximum consecutive degeneration $R$.
\STATE{\textbf{// Domain Similarity Computation}}
\FOR{$q=1\ldots,Q$}
\STATE Call \textbf{Algorithm} \ref{Alg:1} to get the sample set of difference vectors $\mathcal{M}_q$;
\STATE Approximate p.d.f. of the difference vector to get $\hat{Q}_q(D)$ as in Eq. (\ref{eq:kde});
\STATE Generate $\hat{\bm{D}}_q = \left(\begin{array}{c}{\hat{D}_{q,1}}\ {\ldots} \ {\hat{D}_{q,m}} \end{array}\right)^{\top}\in \mathbb{R}^{K\times m}$ from $\hat{Q}_q(D)$;
\STATE Calculate matrix norm 
$
g_q =\frac{1}{m}\left\|\hat{\bm{D}}_q\right\| \in \mathbb{R};
$
\ENDFOR
\STATE{\textbf{// Pre-training in Source Domains}}
\STATE Sort the source domains such that $g_1\geqslant g_2\geqslant\ldots g_Q$ and set $\bm{\theta}_0^{0}=\bm{\theta}^0$;
\FOR{$q = 1,2\ldots,Q$}
\STATE Compute the weight of each source domain by Eq.~\eqref{importance-value}
\STATE Set $\bm{\theta}_{q}^{0}=\bm{\theta}^J_{q-1}$ and $\lambda_q^{0}=\lambda^0$;
\FOR{$j = 0,1,2,\ldots,J$}
\STATE Update the parameter $\boldsymbol{\theta}_{q}^{j+1}$ via Eq.~\eqref{iterate:weight}-\eqref{iterate:lr}
\ENDFOR
\ENDFOR
\STATE{\textbf{// Fine-tuning in Target Domain}}
\STATE Set $\bm{\theta}^{0}_{\mathcal{T}}=\bm{\theta}^{J}_Q$, $\lambda_{\mathcal{T}}^{0}=\lambda_{\mathcal{T}}$, $r=0$ and $j=0$;
\STATE Randomly partition $\hat{\mathcal{T}}$ as $\left\{\mathcal{B}_1,\mathcal{B}_2,\ldots, \mathcal{B}_{\bm{k}}\right\}$;
\WHILE{$r< R$ and $j\leqslant J_\text{target}$}
\STATE Randomly select $\mathcal{B}_j\in \left\{\mathcal{B}_1,\mathcal{B}_2,\ldots, \mathcal{B}_{\bm{k}}\right\}$;
\STATE Compute the learning rate $\lambda_{\mathcal{T}}^j$ by Eq. \eqref{eq:targetlearningrate};
\IF{$\lambda_{\mathcal{T}}^{j}>\lambda_{\mathcal{T}}^{j-1}$}
    \STATE Set $r=r+1$;
\ELSE
    \STATE Set $r=0$;
\ENDIF
    \STATE Update the parameter $\bm{\theta}_{\mathcal{T}}^{j+1}$ via Eq.~\eqref{iterate:target}-\eqref{eq:target} and set $j=j+1$;
\ENDWHILE
\STATE Return $\bm{\theta}_{\mathcal{T}}^{j}$ as the fine-tuned model.
\end{algorithmic}
\label{MSTL}
\end{algorithm}

\section{Experiments} 
\label{sec:exp}
In this section, we conduct numerical experiments to demonstrate the efficacy of our proposed framework. Utilizing time series data collected from motion sensors, we sought to discern and interpret various physical activities, aligning with the daily physical activity monitoring for healthcare applications.

In terms of the dataset, we select the UCI Daily and Sports Activity (DSA) dataset \citep{altun2010human,altun2010comparative,barshan2014recognizing}.
This dataset contains motion sensor data of 19 daily and sports activities carried out by 8 subjects. In particular, participants performed instructed activities while 5 sensors (domains) were placed on the torso, right arm, left arm, right leg, and left leg during data collection. Each sensor captures time series data as a $K=9$ dimensional vector with a length of $T=125$. Each activity comprises 480 time series recordings, summing up to a total of $N=480\times 19$ time series data per domain. In our experiments, we randomly choose data from 6 out of 8 subjects as the training set in each repetition, with the data from the remaining subjects reserved for validation. For data processing, min-max rescaling is applied to each time series dimension, ensuring that all values are confined within the range of $[-1,1]$. This normalization step serves to neutralize the impact of disparities in scale and range between dimensions, thereby enhancing the convergence of stochastic gradient descent during the training phase of the classifiers.

We include several baseline approaches in the experiments as follows: (1) \emph{No Transfer}: This approach does not utilize the time series data in source domains and directly fine-tunes the model in the target domain. (2) \emph{Direct Transfer}: This does not calculate the domain similarities and sets the equal learning rate across all source domains. (3) \emph{No pairing}: This approach pre-trains the model in source domains with the approximated domain distance to the target domain, while the distance between two domains does not take the pairwise structure of the data into consideration. (4) \emph{Freezing}: The freezing method keeps specific layers or weights of a pre-trained model unchanged during the fine-tuning process, allowing the target domain data to update only the unfrozen layers or weights. (5) \emph{Convolutional deep Domain Adaptation model for Time Series data (CoDATS)}: This method applies domain adaptation techniques to align feature distributions \citep{wilson2020multi}. In addition, we also employ different categories of models as the classifier in the experiments, including 
(1) \emph{Long short-term memory networks (LSTM)} \citep{hochreiter1997long}; 
(2) \emph{Encoder} \citep{serra2018towards}; 
(3) \emph{residual neural network (ResNet)} \citep{wang2017time}; and 
(4) \emph{Time series attentional prototype network (TapNet)} \citep{zhang2020tapnet}.

\subsection{General Evaluation}
\label{sec:results}
In this section, we first present the experimental results on classification accuracy, which is quantified using the ratio of correct classification (RCC):
\begin{equation*}
    \operatorname{RCC}=\frac{1}{N_{\text{test} }} \sum_{i=1}^{N_{\text{test} }}\mathbb{I}\left \{ \bm{f}\left ( X_i;\hat{\bm{\theta}}  \right ) =C_i \right \},
\end{equation*}
where $N_{\text{test}}$ denotes the size of the validation set, $X_i$ represents one time series data in the validation set and $C_i$ is the associated label, and $\bm{f}\left ( X;\hat{\bm{\theta}}  \right )$ is the trained classifier. Essentially, RCC measures the alignment between the classifier's output label and the actual ground-truth label across the validation dataset. The entire process is repeated for $I=15$ times. We report both the mean and the standard deviation of RCC across these 15 repetitions.

To align with real application scenarios, such as using wearable sensors to detect falls in vulnerable populations \citep{kavuncuouglu2022investigating,turan2022classification,kocsar2023new}, our experiments focus on binary classification. In each set of experiments, we first randomly select a label as the positive, and the remaining labels are all regarded as the negative. Then the classifier is trained to decide whether the time series is associated with the positive label. Regarding the training, we bootstrap the positive samples (upsampling) so that they have the same number of negative samples. When testing the trained model, we randomly select negative samples (downsampling) to ensure a balance. The experimental results for the dynamic time warping (DTW) metric are included in Table \ref{table:1a}, where the principal number indicates the mean RCC and the value within parentheses represents the standard deviation.
\begin{table*}[ht!]
    \caption{Accuracy of different algorithms with DTW metric on DSA dataset.}
    \label{table:1a}
    \centering
    \begin{tabular}{c|cccc}
        \hline
        \hline
        Algorithm & LSTM & Encoder & ResNet & TapNet\\
        \hline
        DTW-Paired (ours) & $\mathbf{.9722 (\pm .0104)}$ & $\mathbf{.9655 (\pm .0126)}$ & $.9524 (\pm .0155)$ &  $\mathbf{.9726 (\pm .0122)}$ \\
        No Transfer & $.8451 (\pm .0267)$ & $.7632 (\pm .0062)$ & $.6164 (\pm .0204)$ &  $.7352 (\pm .0114)$ \\
        Direct Transfer & $.8729 (\pm .0109)$ & $.8856 (\pm .0134)$ & $.9255 (\pm .0134)$ & $.8331 (\pm .0374)$ \\
        No pairing & $.9184 (\pm .0214)$ & $.9265 (\pm .0124)$ & $.9310 (\pm .0102)$ & $.8977 (\pm .0212)$ \\
        Freezing  & - & $.9112 (\pm .0137)$ & $\mathbf{.9655 (\pm .0032)}$ & $.9271 (\pm .0134)$ \\
        CoDATS  & $.9392 (\pm .0054)$ & $.9627 (\pm .0185)$ & $.9292 (\pm .0157)$ & $.9622 (\pm .0153)$ \\
        \hline
    \end{tabular}
\end{table*}
The findings from our results offer several key insights: Our proposed transfer learning framework with DTW metric consistently achieves the highest classification accuracy when using LSTM, Encoder and TapNet classifiers. With the ResNet classifier, 
its performance is ranked second and comparable to the best. In addition, 
\emph{No Transfer} 
obtains the least classification accuracy, underlining the integral role of transfer learning with multiple sources in boosting classification performance. While \emph{Direct Transfer} and \emph{No Pairing}
do improve the performance, they lag considerably behind our proposed method. This underscores the potential of the inherent structure of data, suggesting that disregarding it can dilute the quality of results. The state-to-art CoDATS method achieves slightly worse results for all classifiers, which further validates the efficiency of our method. Among various transfer learning technologies and classifiers, our framework with the TapNet model achieves the best performance. We include additional numerical experiments in Appendix, which also demonstrates the priority of our framework.


\subsection{In-depth Evaluation}
In this section, we delve deeper into the performance evaluation of our proposed approach. Specifically, we first compare the results with different time series distance metric. Then we conduct experiments to evaluate whether the order of source domains in the pre-training phase affects the performance of our approach. Lastly, we impose noise to the time series data when testing the algorithms.

\subsubsection{Sensitivity on Distance Metric}
\label{sec:timedistance}
In the previous experiment, we select the DTW distance as a representative of the time series distance metric. We further include the experimental results with another two distances: (1) Euclidean distance and (2) the Bag-of-SFA Symbols (BOSS) algorithm \citep{schafer2015boss}. We present the experimental results in Table \ref{table:distance}. The results indicate that the adaptive transfer learning approach with DTW consistently outperforms other metrics. This can be attributed to the capability of DTW to manage non-linear alignment between time series. It excels at capturing similarities even when patterns in the data have different rates of progression or occur in different phases. However, the experiments with the Euclidean distance deliver satisfactory results, meanwhile calculating the Euclidean distance is more efficient than DTW.

\begin{table*}[htbp]
    \caption{Accuracy of adaptive transfer learning with different time series distance metrics on DSA dataset.}
    \label{table:distance}
    \centering
    \begin{tabular}{c|cccc}
        \hline
        \hline
        Distance metric & LSTM & Encoder & ResNet & TapNet\\
        \hline
        DTW & $\mathbf{.9722 (\pm .0104)}$ & $\mathbf{.9655 (\pm .0126)}$ & $\mathbf{.9524 (\pm .0104)}$ & $\mathbf{.9726 (\pm .0122)}$\\
        Euclidean & $.9268 (\pm .0034)$ & $.9288 (\pm .0206)$ & $.9254 (\pm .0204)$ & $.9432 (\pm .0206)$ \\
        BOSS & $.9310 (\pm .0116)$ & $.9492 (\pm .0105)$ & $.9492 (\pm .0105)$ & $.9421 (\pm .0221)$ \\
        \hline
    \end{tabular}
\end{table*}

\subsubsection{Order of Source Domains in the Pre-training Phase}
\label{sec:order}
Recall that, in the pre-training procedure of our proposed framework, we sort and renumber all the source domains in descending order based on the associated IPD. We also conduct experiments where the order of the source domains is randomly determined. We present the results in Table \ref{table:order}.
\begin{table*}[ht!]
    \caption{Accuracy of adaptive transfer learning with different orders on DSA dataset.}
    \label{table:order}
    \centering
    \begin{tabular}{c|cccc}
        \hline
        \hline
        Order & LSTM & Encoder & ResNet & TapNet\\
        \hline
        Sorted & $\mathbf{.9722 (\pm .0104)}$ & $\mathbf{.9655 (\pm .0126)}$ & $\mathbf{.9524 (\pm .0104)}$ & $\mathbf{.9726 (\pm .0122)}$\\
        Random & $.9465 (\pm .0315)$ & $.9232 (\pm .0115)$ & $.9155 (\pm .0434)$ & $.9552 (\pm .0458)$ \\
        \hline
    \end{tabular}
\end{table*}
We have the following insights. First, when we employ a sorted order for the source domains in Algorithm \ref{MSTL}, there is a noticeable improvement in the RCC compared to a random order. This enhancement can be attributed to the process wherein the pre-trained model, in a sorted order, is predominantly updated and influenced by data from source domains that align more closely with the target domain. Second, when adaptive transfer learning uses a random order for source domains, it tends to display a 
higher standard deviation in RCC. This is because a random order brings inherent unpredictability during the pre-training phase, leading to more uncertainty. This inconsistency carries through, affecting the performance of the trained model. Lastly, despite the challenges posed by randomness, the adaptive transfer learning framework manages to deliver satisfactory classification results even with a random order of source domains owing to the learning rate tailored for each source domain. Since this rate factors in the similarity between the source and target domain, the overall learning process remains relatively stable to the specific sequence of source domains during pre-training.

\subsubsection{Noise Injection}
\label{sec:noise}
\begin{figure}[ht]
\centering
\begin{minipage}[t]{0.48\textwidth}
\centering
\includegraphics[width=\textwidth]{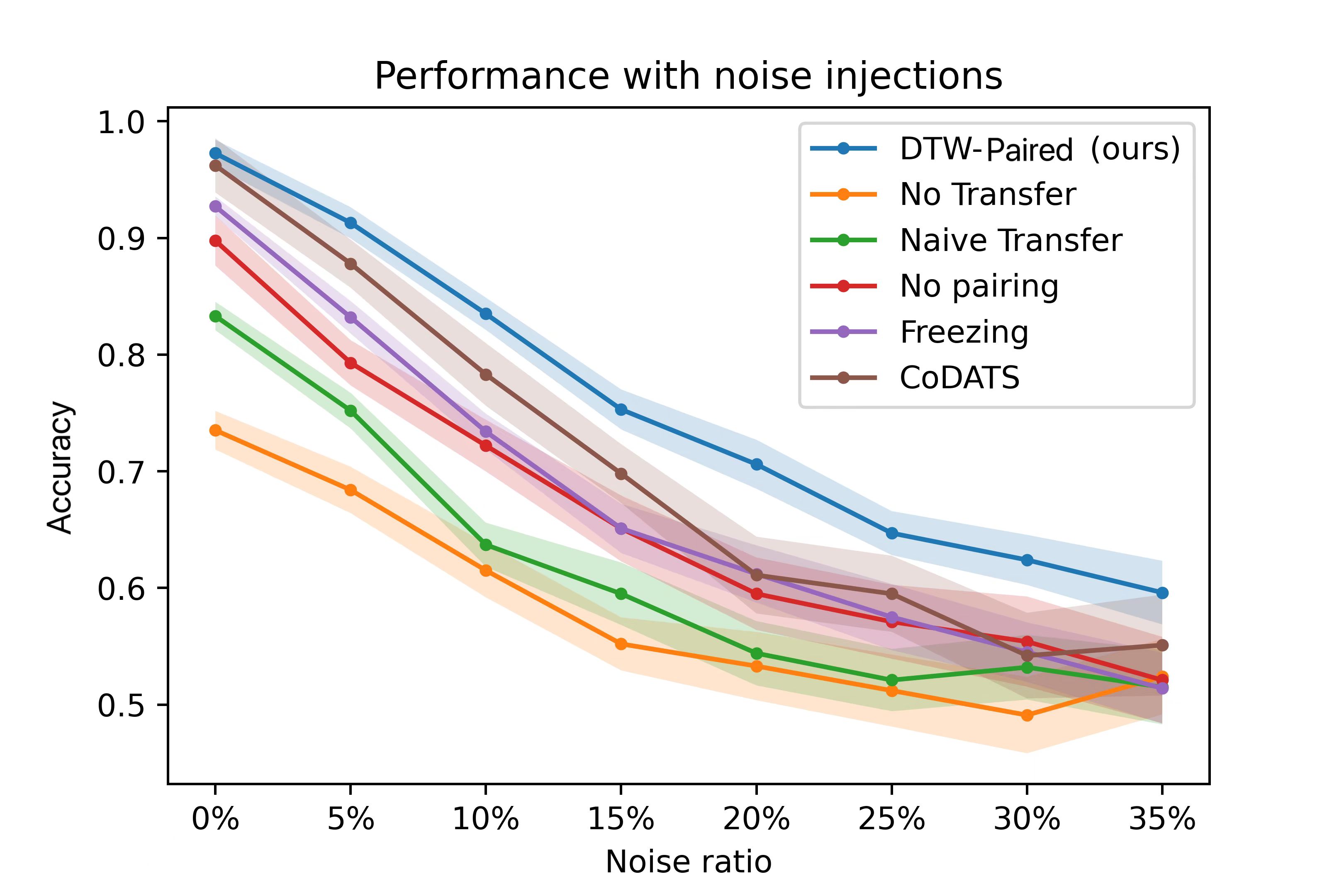}
\end{minipage}
\begin{minipage}[t]{0.48\textwidth}
\centering
\includegraphics[width=\textwidth]{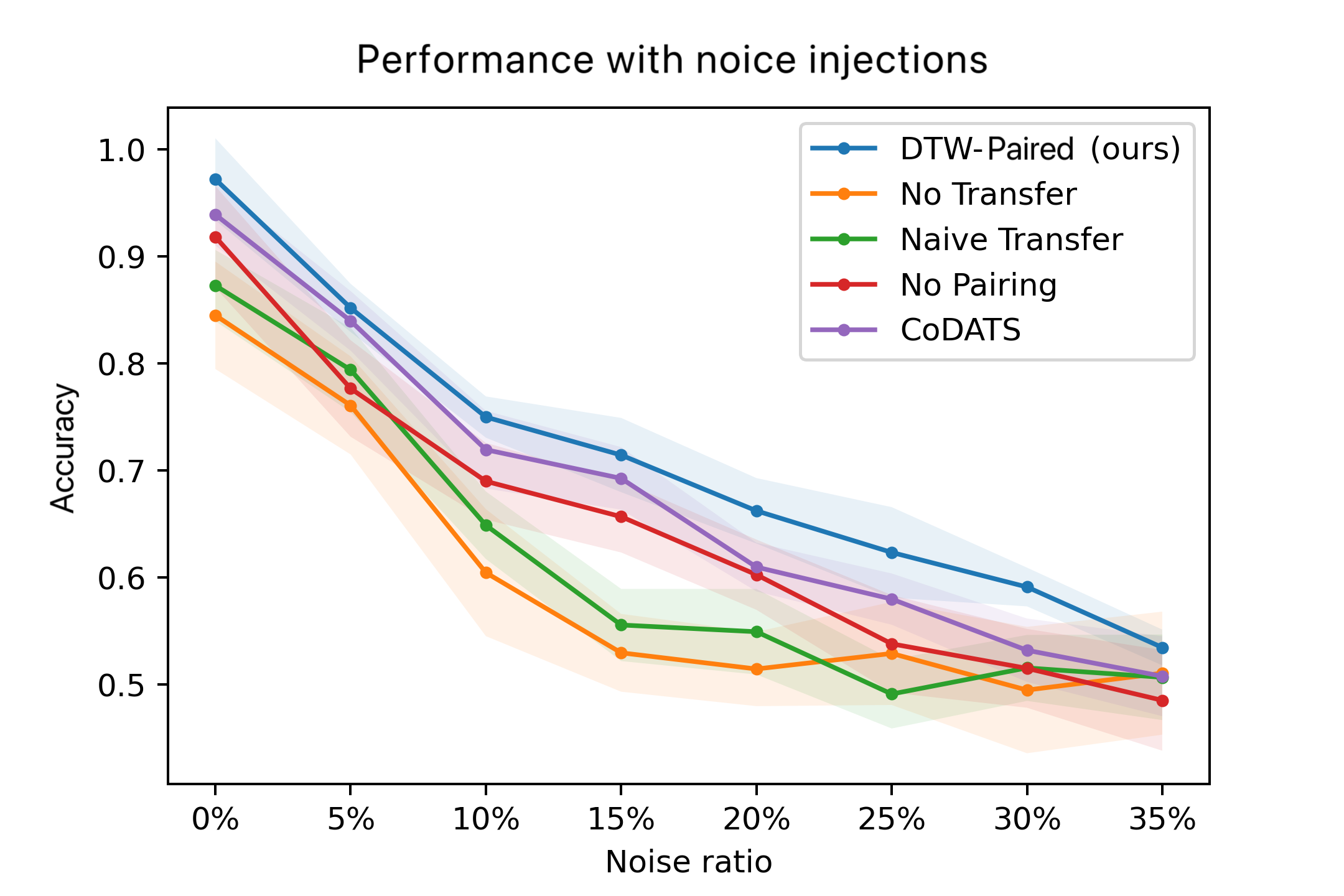}
\end{minipage}
\caption{RCC of time series classification approaches across different ratios of noise. The standard deviation of RCC is indicated by the shadow along the line.}\label{fig:noise}
\end{figure}

In real applications, time series collected by wearable motion sensors are frequently susceptible to noise from the data collection process. This contrasts with data from controlled laboratory settings, which often offer cleaner readings \citep{rubin2023denoising}. As such, the trained classifier using the laboratory data is supposed to be robust against the noise in the input dataset. Therefore, to replicate real-world conditions more accurately, we introduce synthetic Gaussian noise $\mathcal{N}\left(\bm{0},0.02\operatorname{Diag}\left\{x_t\right\}\right)$ to some 
of the input time series data $X$.
Our objective is to evaluate the robustness of the trained classifier against the presence of input noise in the context of wearable motion sensor data. As shown in Figure \ref{fig:noise}, we present the results using both TapNet and LSTM classifiers, factoring in different ratios of the timestamps that are injected with noise, which indicates the robustness of our proposed transfer learning framework.

The experimental results in Figure \ref{fig:noise} indicate that although the input data noise affects all transfer learning methods, our framework achieves the highest accuracy across different noise ratios. In conclusion, our proposed approach achieves robustness against the noise in the input time series data, which is attributed to the utilization of the smooth bootstrap method when calculating the proposed metric that quantifies the domain similarities.

\section{Discussion} 
\label{sec:conclusion}
We develop a transfer learning framework for multi-source time series classification, aiming to enhance daily physical monitoring in healthcare applications (e.g., rehabilitation) using data from multiple motion sensors in professional laboratories. We propose a metric to quantify domain similarities that account for the pairwise structure of time series. The framework then pre-trains a classifier on source domains and fine-tunes it on the target domain, where the degree of knowledge transfer is determined by the proposed metric. Our experimental results show the superiority of our approach, achieving higher classification accuracy and robustness against input data noise compared to existing methods. Thus, the application of transfer learning enhances daily physical monitoring compared to traditional methods that are based on single-source time series data, offering improved health monitoring, diagnosis, and intervention.

Regarding future work, our framework goes beyond daily physical activity monitoring, demonstrating adaptability to a broader range of healthcare assessments such as heart and lung function \citep{mcdermott2021comprehensive}. Our framework compensates for the limited scope of data collected by wearable ECG monitors and simple at-home spirometry devices compared to their professional counterparts \citep{raghu2022data,spathis2021self}. This adaptability, crucial for enhancing data interpretation in daily life, signifies the framework's potential impact on healthcare. To facilitate this, future enhancements focus on accommodating diverse data forms, including images and audio, within healthcare applications \citep{xu2023multi,su2024large}. By adapting our transfer learning framework for multi-modal data integration and developing methodologies for cross-domain comparisons \citep{liu2023adaptive}, we aim to facilitate more comprehensive and accurate patient care. This approach includes exploring advanced feature extraction technologies and new similarity metrics, paving the way for a more versatile and effective application of the framework in leveraging varied healthcare data.

\newpage

\bibliographystyle{informs2014}

\bibliography{references}

\newpage
\begin{APPENDIX}{}
\section{Fine-tuning in Target Domain}
\label{sec:43}
After completing the pre-training procedure, we obtain the pre-trained model parameter $\bm{\theta}_Q^{J}$ from all source domains. The last step of the adaptive transfer learning involves fine-tuning the model based on using the target domain data $\hat{\mathcal{T}} =\left\{X_{\mathcal{T},n}\right\}_{n=1}^N$ and associated labels $\left\{C_n\right\}_{n=1}^N$. The classifier's parameters are initialized with the pre-trained model parameter, denoted as $\boldsymbol{\theta}^0_\mathcal{T}=\boldsymbol{\theta}_Q^J$. For fine-tuning in the target domain, we use the mini-batch gradient descent method combined with $\bm{k}$-fold cross-validation. The target domain dataset $\hat{\mathcal{T}}$ is randomly partitioned into $\bm{k}$ equal-sized, disjoint subsets as $\left\{\mathcal{B}_1,\mathcal{B}_2,\ldots, \mathcal{B}_{\bm{k}}\right\}$. During each learning epoch, we randomly select a subset $\mathcal{B}_i, i\in \bm{k}$ with equal probability as the validation set, while the remaining $\bm{k}-1$ subsets are used as the training set. In this way, the model parameter $\bm{\theta}_\mathcal{\mathcal{T}}^{j}$ is updated in the $j$-th learning epoch as
\begin{equation}\label{iterate:target}\boldsymbol{\theta}_{\mathcal{T}}^{j+1}=\boldsymbol{\theta}_{\mathcal{T}}^j-\lambda_{\mathcal{T}}^j \nabla_{\boldsymbol{\theta}} \mathcal{J}_{\mathcal{T}}\left(\boldsymbol{\theta}_{\mathcal{T}}^j\right),
\end{equation}

Here, $\mathcal{J}_{\mathcal{T}}\left(\boldsymbol{\theta}_{\mathcal{T}}^{j}\right)$ is the loss function given by
\begin{equation}\label{eq:target}
\mathcal{J}_{\mathcal{T}}\left(\boldsymbol{\theta}_{\mathcal{T}}^{j}\right)=\mathbb{E} _{\{\hat{\mathcal{T}}\setminus \mathcal{B}_j\}\cup\mathcal{C}_j}\mathcal{L}\left( \boldsymbol{\theta}_{\mathcal{T}}^{j}\right).
\end{equation}
where $\{\hat{\mathcal{T}}\setminus \mathcal{B}_j\}$ is the training set in the $j$-th epoch and $\mathcal{C}_j$ denotes the set of labels associated with the training set. The learning rate in the $j$-th epoch is determined as follows:
\begin{equation}
\label{eq:targetlearningrate}
\lambda_{\mathcal{T}}^{j}=\left(1-\mathbb{E} _{\mathcal{B}_j\cup\mathcal{C}'_j}\mathcal{L}\left( \boldsymbol{\theta}_{\mathcal{T}}^{j}\right)\right)\lambda_{\mathcal{T}},
\end{equation}
where $\mathcal{B}_j$ is the selected validation set, $\mathcal{C}'_j$ is the associated set of labels and $\lambda_\mathcal{T}$ is the prescribed baseline learning rate in the target domain. The performance of the current model is evaluated with the validation set, and the learning rate for the next epoch is adjusted accordingly, decreasing if the current model exhibits good performance. The learning procedure stops if there is performance degeneration on the validation set for $R$ consecutive epochs or if the learning epoch has been repeated for $J_\text{target}$ times. Here $R$ and $J_\text{target}$ are prescribed hyperparameters. The complete procedure of the proposed transfer learning with multi-source time series data is summarized in Algorithm \ref{MSTL}.


\subsection{Domain Shift}
In addition to fine-tuning the classifier with the target domain data, we also include a discussion on the additional domain shift here. That is, although the target domain aligns with the wearable motion sensor in daily use regarding the body part, the data collected in laboratory settings are under a more controllable environment. In this way, there is an additional domain shift between the distribution of the target domain data and that of daily collected data. 

In the experiments presented in the main text, we, therefore, consider imposing additional noise to the time series when testing the trained classifier. This procedure imitates the real environment with more noise. Experimental results indicate that our transfer learning framework is more robust to the noise compared with the baseline approaches, owing to the calculation of $g_q$. Moreover, our framework can manage broader domain shifts through domain adaptation techniques if more labeled time series data is gathered in daily use. Test Time Adaptation (TTA) is one feasible approach, allowing the model to adapt during the testing phase to bridge the gap between the distributions of the training and testing data. We here summarize a general procedure of implementing TTA that is applicable to our framework:
\begin{enumerate}
    \item \textbf{Identifying Domain Shift:} Before adjusting the model (classifier), the distribution shift is required to be identified. Methods include: 1) applying statistical tests (e.g., Chi-square test) to compare the distribution of features in the training data and the newly collected data and 2) monitoring the performance of the classifier on the new data. A significant drop in performance metrics (accuracy, precision, recall) might be attributed to a domain shift.

    \item \textbf{Data Processing:} If the distribution shift is identified, the classifier is then required to be adjusted. It should be guaranteed that the new data are processed and normalized in the same way as the training data so that the classifier can learn from the new data.

    \item \textbf{Model Updating:} The model is updated by fine-tuning with new data. Specifically, the learning rate of fine-tuning is suggested to be small and decrease, ensuring a conservative updating.

    \item \textbf{Rolling Validation:} Implement a rolling validation procedure where the classifier is periodically validated using a recent subset of the data. This continuous validation facilitates ongoing observation of the adaptation process.

    \item \textbf{Continuous Monitoring and Adaptation:} Continuously monitor the model's performance on new data. When there is a domain shift, iterate previous steps to adjust the classifier.

\end{enumerate}

We refer to \cite{wang2020tent,ma2022test,lee2022diversify} for detailed procedures of TTA. Since the open-access datasets used in our experiments are collected in a stationary manner and do not exhibit domain shifts, we do not include TTA in our experiments. We defer the specific integration of TTA with our framework to future work.

\section{Experiment Details}
\label{sec:detail}
In the experiments, regarding our proposed adaptive transfer learning framework described in Algorithm \ref{MSTL}, we set:
\begin{enumerate}
    \item The initial learning rate $\lambda^0=5\times 10 ^{-4}$;
    \item The number of learning epochs in each source domain $J=50$;
    \item The number of learning epochs in the target domain $J_{\text{target}}=100$;
    \item The value of the model parameter is initiated by randomly sampling each weight parameter of the neural network from $\operatorname{Uniform}[0,1]$ and each bias parameter as $0$;
    \item The number of partitions $\bm{k}=10$;
    \item The baseline learning rate in the target domain $\lambda_{\mathcal{T}}=1\times 10^{-3}$;
    \item The number of the maximum consecutive degeneration $R=5$.
\end{enumerate}

In terms of the neural network models employed as the classifiers, we consider (1) Long short-term memory networks (LSTM); (2) Encoder; (3) Residual neural network (ResNet); and (4) Time series attentional prototype network (TapNet). We briefly describe the employed models as follows.
\begin{enumerate}
    \item LSTM uses three gates to control the information flow of a sequence of data, which can capture the hidden patterns of input sequences \citep{hochreiter1997long}. The exact implementation follows \url{https://pytorch.org/docs/stable/generated/torch.nn.LSTM.html?highlight=lstm#torch.nn.LSTM}. We also note that the freezing technology for transfer learning is not applicable to the LSTM models due to their sequential nature, which makes learned features highly interconnected and task-specific. LSTMs also have limited depth, reducing the chances of hierarchical feature learning that enables transfer learning with layer freezing in deep neural networks. Instead, alternative transfer learning strategies are more suitable for LSTM models.
    
    \item Encoder applies deep neural networks compressing the raw input sequences into a low-dimensional representation, and make predictions and classifications directly based on the encoded variable. We refer to \cite{serra2018towards} for reference and the implementation of Encoders can be found in \url{https://github.com/sktime/sktime-dl/blob/master/sktime_dl/regression/_encoder.py}.
    \item ResNet introduces the residual connection in neural networks, which can avoids gradient vanishing and information loss in learning the pattern of a sequences. We refer to \cite{wang2017time} for reference and \url{https://github.com/sktime/sktime-dl/blob/master/sktime_dl/regression/_resnet.py} for implementation.
    \item TapNet applies temporal attention mechanism to learn the importance of different timesteps of a sequence. Exact implementation follows \cite{zhang2020tapnet} and \url{https://www.sktime.net/en/latest/api_reference/auto_generated/sktime.classification.deep_learning.TapNetClassifier.html?highlight=tapnet}.
\end{enumerate}

All the experiments were run by Python 3.8 and Pytorch on a server with two 32-Core AMD Ryzen Threadripper PRO 3975WX processors and three NVIDIA RTX A6000 GPUs.

\section{Additional Experiments}
We present additional numerical experiments in this Section.

\subsection{Influential Source Domains}
In this section, we present the two most influential source domains for each target domain in \textbf{Table \ref{table:inf}}, based on the DSA dataset. Given a specific target domain, the most influential domains are determined by the calculated $g_q$. Specifically, the smaller $g_q$ is, the more similarities there are between the two domains, indicating that the source domain is more influential.

\begin{table*}[ht!]
    \caption{The two most influential source domains for each target domain.}
    \label{table:inf}
    \centering
    \begin{tabular}{c|cc}
        \hline
        \hline
        Target Domain & \thead{Most Influential\\Source Domain} & \thead{2nd Most Influential \\
        Source Domain}\\
        \hline
        Torso & Left Leg & Right Leg  \\
        Right Arm &Left Arm & Right Leg  \\
        Left Arm & Right Arm& Left Leg  \\
        Left Leg & Right Leg & Left Arm  \\
        Right Leg & Left Leg & Right Arm  \\
       
        \hline
    \end{tabular}
\end{table*}

In addition, we also consider pre-training the time series classifiers using the two most influential source domains, instead of across all source domains. We present the numerical results in \textbf{Table \ref{table:1i}}. Compared to pre-training across all source domains, excluding the least influential source domains leads to a decrease in classification accuracy but also slightly reduces model uncertainty, as indicated by the standard deviation of the classification accuracy. Moreover, focusing pre-training on the most influential source domains results in time savings during training. Thus, whether to select the most influential domains for pre-training depends on the requirements of the applications. We also note that adaptive transfer learning from the two most influential source domains using our framework still outperforms existing algorithms.

\begin{table*}[ht!]
    \caption{Accuracy of different source domain selections on DSA dataset.}
    \label{table:1i}
    \centering
    \begin{tabular}{c|cccc}
        \hline
        \hline
        Source Domain Selection & LSTM & Encoder & ResNet & TapNet\\
        \hline
        All Source Domains & $\mathbf{.9722 (\pm .0104)}$ & $\mathbf{.9655 (\pm .0126)}$ & $\mathbf{.9524} (\pm .0155)$ &  $\mathbf{.9726 (\pm .0122)}$ \\
        \thead{Two Most Influential\\Source Domains} & $.9533 (\pm .0087)$ & $.8911 (\pm .0092)$ & $.9124 (\pm .0118)$ &  $.9324 (\pm .0096)$ \\
        \hline
    \end{tabular}
\end{table*}

\subsection{Additional Experiments on RSS Data Set}
In this section, we present the experimental results based on another data set that contains time series data collected by multiple motion sensors. Specifically, we select the data set `` Indoor User Movement Prediction from RSS data Data Set''.
The data set can be used for a binary classification task consisting of predicting the pattern of user movements from time series generated by a Wireless Sensor Network (WSN). Input data contains temporal streams of radio signal strength (RSS) measured between the nodes of a WSN, comprising 5 sensors. For the given dataset, RSS signals have been re-scaled to the interval $[-1,1]$, singly on the set of traces collected from each anchor. Target data consists of a class label indicating whether the user's trajectory will lead to a change in the spatial context (i.e. a room change) or not.

The experimental settings are the same as the experiments on the DSA data set and the experimental results are included in Table \ref{table:2a}. The results provide insights as follows. First, in this set of experiments, our proposed adaptive transfer learning framework achieves the best performance across different selections of classifiers. Second, directly fine-tuning the classifier in the target domain without transfer learning from source domains achieves acceptable results. Consequently, knowledge from source domains without appropriate methodology does not always enhance the classification performance, which is different from the experimental results in Section \ref{sec:exp}. This may be attributed to the fact that the task here is a binary classification problem, which is simpler, and therefore, the target domain data provides enough information to facilitate the task. Lastly, our proposed approach with the utilization of TapNet achieves the best classification performance across different transfer learning methodologies and employed classifiers, which is the same as the experimental results in Section \ref{sec:exp}. 

\begin{table*}[ht!]
    \caption{Accuracy of different algorithms with DTW metric on dataset `Indoor User Movement Prediction from RSS'.}
    \label{table:2a}
    \centering
    \begin{tabular}{c|cccc}
        \hline
        \hline
        Algorithm & LSTM & Encoder & ResNet & TapNet\\
        \hline
        DTW-Paired (ours) & $\mathbf{.9722 (\pm .0075)}$ & $\mathbf{.9865 (\pm .086)}$ & $\mathbf{. 9923 (\pm .0044)}$ & $\mathbf{.9926 (\pm .0004)}$ \\
        No Transfer & $.9138 (\pm .0312)$ & $.8742 (\pm .0072)$ & $.9704 (\pm .0109)$ & $.9694 (\pm .0052)$ \\
        Direct Transfer & $.9428 (\pm .0230)$ & $.9256 (\pm .094)$ & $.9255 (\pm .0134)$ & $.9744 (\pm .0134)$ \\
        No pairing & $.9514 (\pm .0064)$ & $.9566 (\pm .084)$ & $.9310 (\pm .0102)$ & $.9585 (\pm .0262)$ \\
        Freezing & - & $.9612 (\pm .0137)$ & $.9245 (\pm .0032)$ & $.9474 (\pm .0064)$ \\
        CoDATS & $.9673 (\pm .0071)$ & $.9797 (\pm .0078)$ & $.9899 (\pm .0102)$ & $.9612 (\pm .0182)$ \\
        \hline
    \end{tabular}
\end{table*}

\end{APPENDIX}

\end{document}